\documentclass[sigconf]{acmart}

\usepackage{graphicx}
\usepackage{amsmath}
\usepackage{booktabs}
\usepackage{subfloat}
\usepackage{subcaption}
\usepackage{multirow}
\usepackage{wrapfig}
\usepackage[T1]{fontenc}
\usepackage{xcolor}
\usepackage[normalem]{ulem}
\usepackage[linesnumbered,lined,boxed,commentsnumbered,ruled]{algorithm2e}
\usepackage{algpseudocode}
\usepackage{balance}

\copyrightyear{2023}
\acmYear{2023}
\setcopyright{acmlicensed}\acmConference[MM '23]{Proceedings of the 31st
ACM International Conference on Multimedia}{October 29-November 3,
2023}{Ottawa, ON, Canada}
\acmBooktitle{Proceedings of the 31st ACM International Conference on
Multimedia (MM '23), October 29-November 3, 2023, Ottawa, ON, Canada}
\acmPrice{15.00}
\acmDOI{10.1145/3581783.3612319}
\acmISBN{979-8-4007-0108-5/23/10}

\begin{document}
\title{MetaFBP: Learning to Learn High-Order Predictor for Personalized Facial Beauty Prediction}

\author{Luojun Lin}
\affiliation{%
  \institution{Fuzhou University}
  \city{Fuzhou}
  \country{China}}
\email{linluojun2009@126.com}
\orcid{0000-0002-1141-2487}

\author{Zhifeng Shen}
\affiliation{%
  \institution{Fuzhou University}
  \city{Fuzhou}
  \country{China}}
\email{shen_zhifeng@outlook.com}
\orcid{0000-0002-2215-4479}

\author{Jia-Li Yin}
\affiliation{%
  \institution{Fuzhou University}
  \city{Fuzhou}
  \country{China}}
\email{jlyin@fzu.edu.cn}
\orcid{0000-0002-8087-9769}

\author{Qipeng Liu}
\affiliation{%
  \institution{Fuzhou University}
  \city{Fuzhou}
  \country{China}}
\email{cometq@foxmail.com}
\orcid{0000-0002-0903-7171}

\author{Yuanlong Yu}
\authornote{Y. Yu and W. Chen are corresponding authors.}
\affiliation{%
  \institution{Fuzhou University}
  \city{Fuzhou}
  \country{China}}
\email{yu.yuanlong@fzu.edu.cn}
\orcid{0000-0002-2112-6214}

\author{Weijie Chen}
\authornotemark[1]
\affiliation{%
  \institution{Zhejiang University \& Hikvision Research Institute}
  \city{Hangzhou}
  \country{China}}
\email{chenweijie@zju.edu.cn}
\orcid{0000-0001-5508-473X}

\renewcommand{\shortauthors}{Luojun Lin et al.}

\begin{abstract}
Predicting individual aesthetic preferences holds significant practical applications and academic implications for human society. However, existing studies mainly focus on learning and predicting the commonality of facial attractiveness, with little attention given to \emph{Personalized Facial Beauty Prediction} (PFBP). PFBP aims to develop a machine that can adapt to individual aesthetic preferences with only a few images rated by each user. In this paper, we formulate this task from a meta-learning perspective that each user corresponds to a meta-task. To address such PFBP task, we draw inspiration from the human aesthetic mechanism that visual aesthetics in society follows a Gaussian distribution, which motivates us to disentangle user preferences into a commonality and an individuality part. To this end, we propose a novel MetaFBP framework, in which we devise a universal feature extractor to capture the aesthetic commonality and then optimize to adapt the aesthetic individuality by shifting the decision boundary of the predictor via a meta-learning mechanism. 
Unlike conventional meta-learning methods that may struggle with slow adaptation or overfitting to tiny support sets, we propose a novel approach that optimizes a high-order predictor for fast adaptation. In order to validate the performance of the proposed method, we build several PFBP benchmarks by using existing facial beauty prediction datasets rated by numerous users. Extensive experiments on these benchmarks demonstrate the effectiveness of the proposed MetaFBP method.
\end{abstract}



\maketitle

\section{Introduction}
\begin{figure}[!t]
\centering
\includegraphics[width=1.0\columnwidth]{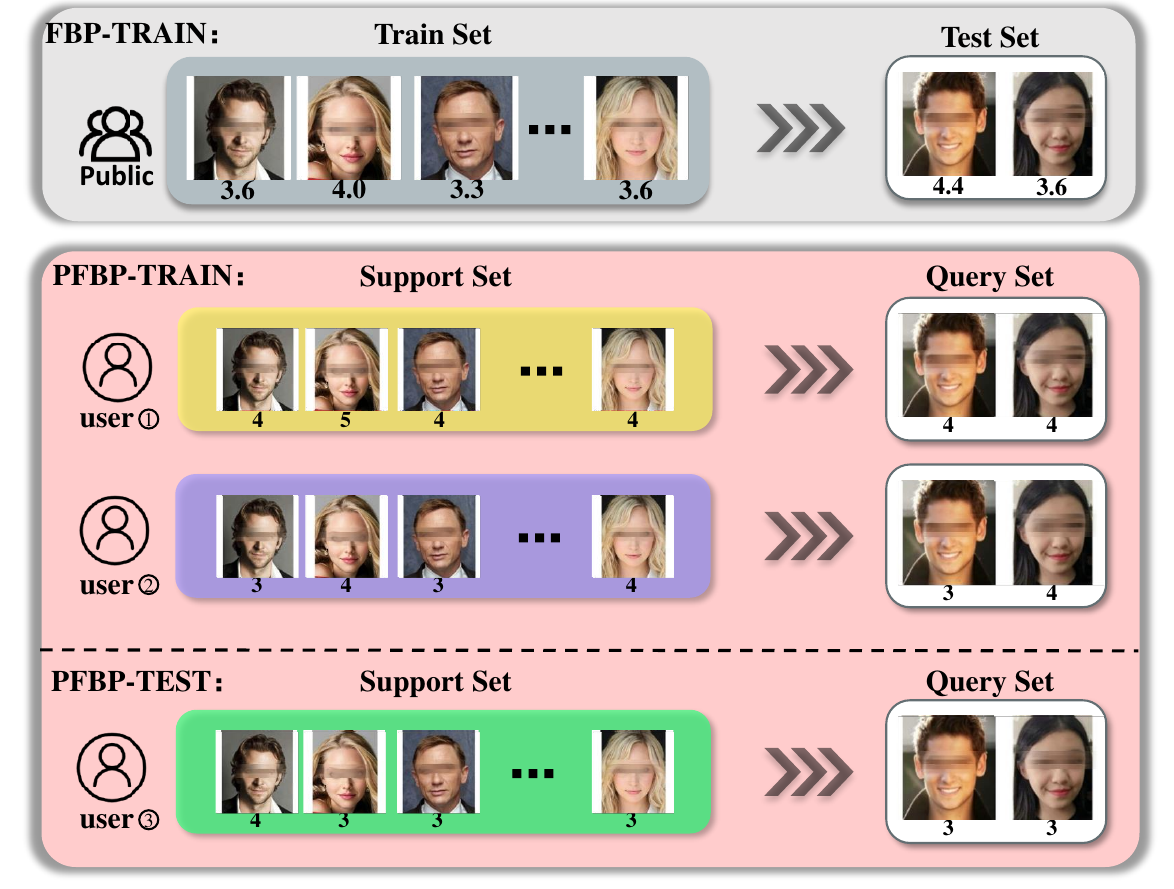}
\caption{The difference between FBP and PFBP. The conventional FBP only gives an average beauty score of the public for a facial image, but PFBP provides different beauty scores for each facial image according to user preference. Note that each face has been locally pixelated for privacy protection.}
\label{fig:intro_fig1}
\end{figure}

Facial Beauty Prediction (FBP) has allured many research interests in recent years. Most studies aim to develop models that can accurately judge facial beauty in line with the average aesthetic preferences of a large population of users, using average ratings as the supervision signals for model learning~\cite{lin2019attribute, chen2016data, lin2019regression}. However, these studies merely focus on the commonality and overlooks the highly subjective nature of human aesthetic perception, as illustrated in Figure \ref{fig:intro_fig1}, where each facial image is rated by different users with various attractiveness scores. Therefore, the subjective nature of facial aesthetics should be taken into account to develop more accurate predictions of facial attractiveness for different users.

To fill the current gap in this area, we delve into \emph{Personalized Facial Beauty Prediction} (PFBP) in this paper. \emph{The objective of this task is to make consistent aesthetic judgments with a specific user by previously requiring the user to rate a few facial images.} A highly-performing PFBP model has significant practical applications in various online systems, such as social recommendation systems or make-up recommendation systems~\cite{chen2023customized}. As shown in Figure \ref{fig:intro_fig1}, the recommendation system requires each user to label a few facial images to adapt the PFBP model so that it can quickly capture the aesthetic preference and then send the top-ranked recommended faces from the image gallery to the target user. From this perspective, PFBP is expected to possess a fast adaptation ability for each user preference with limited labeled data.

Considering the user-adaptive and data-limited properties of the PFBP task, it intuitively motivates us to reformulate PFBP from a few-shot learning perspective~\cite{finn2017model}. Specifically, each individual user corresponds to a meta-task consisting of a support set and a query set.
The training and evaluation of the PFBP model follow the meta-training and meta-testing stages in few-shot learning. Nevertheless, there are still two main differences between PFBP and conventional few-shot learning tasks:
\begin{enumerate}
    \item[1)] In conventional few-shot learning tasks, the categories of each meta-task are different. The training goal is to quickly adapt the model from base to novel categories with limited training data. Unlike them, the categories of each meta-task in PFBP are fixed. The range of attractiveness score is shared among different meta-tasks. The training objective is not to adapt to novel categories but to novel users with specific aesthetic preferences.
    \item[2)] In conventional few-shot learning tasks, the labels of images are fixed across meta-tasks, \emph{e.g.}, the label of a cat image always belongs to ``cat'' and cannot be changed to other categories. However, in PFBP, the attractiveness score of a facial image will change across different meta-tasks because users have different aesthetic preferences and thus give different ratings to the same image.
\end{enumerate}

\begin{figure}[!t]
\centering
\includegraphics[width=1.0\columnwidth]{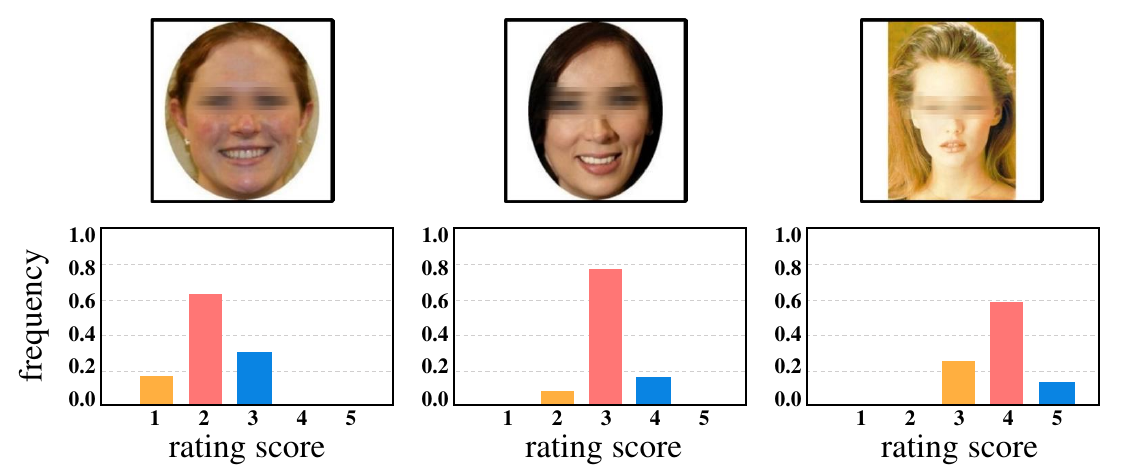}
\caption{The rating distributions of three randomly-selected images rated by a population of volunteers. The aesthetic preference roughly follows a Gaussian distribution.}
\label{fig:intro_fig2}
\end{figure}

The main challenge in PFBP is the subjective nature of user ratings, \emph{i.e.}, the changeability of image labels across meta-tasks, which never occurs in previous image recognition tasks. It urges a rather strong adaptability of the PFBP model that can forget the image labels seen in previous meta-tasks and adapt to current meta-task quickly with limited labeled data. Despite the variability of ratings for facial images, we have observed that the attractiveness score of an image rated by a population of users tends to follow a Gaussian distribution, as shown in Figure \ref{fig:intro_fig2}. That is, the population aesthetic tends to be consistent while the personalized preference is fluctuated around the population aesthetic.
This phenomenon can be attributed to the objective part of human aesthetic perception, namely aesthetic commonality, which plays a crucial role in working alongside the subjective part, aka aesthetic personality.

\begin{figure}[!t]
\centering
\includegraphics[width=1.0\columnwidth]{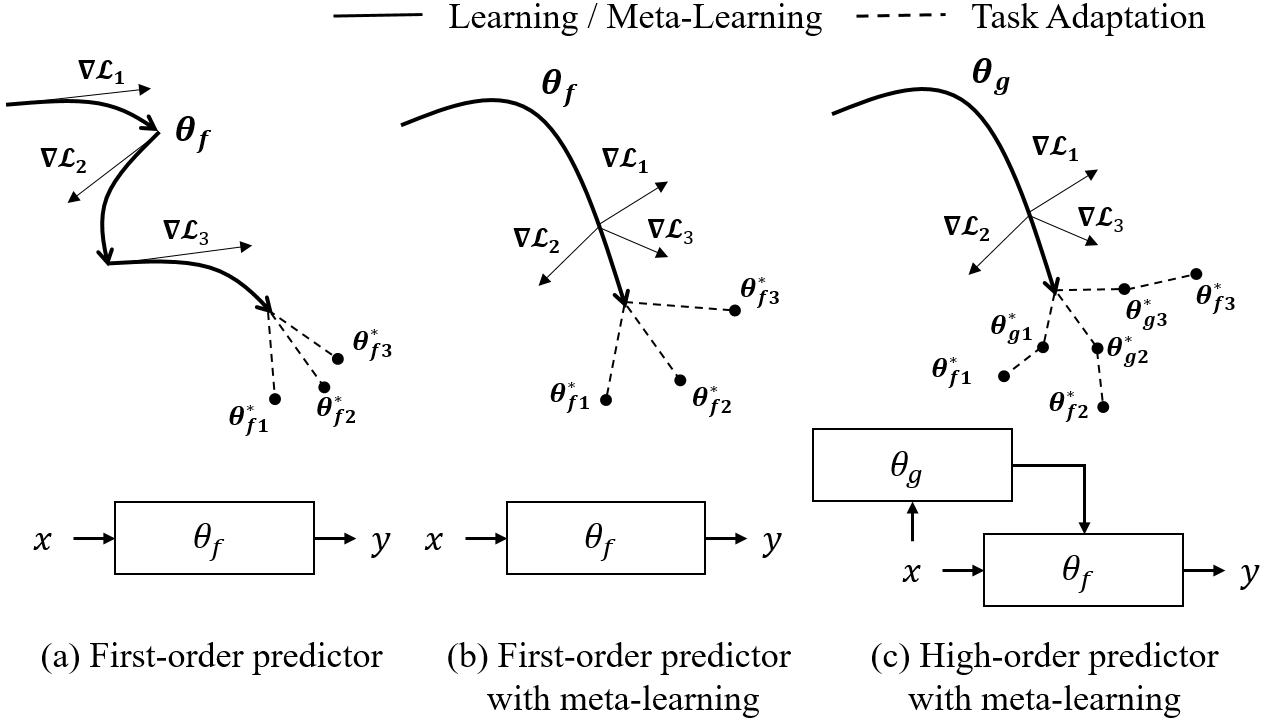}
\caption{The comparison of different learning paradigms. (a) Common learning paradigm. (b) Conventional meta-learning paradigm. (c) Our proposed paradigm that involves learning to learn high-order predictor for fast adaptation.}
\label{fig:method}
\end{figure}

Motivated by the observations, we propose to disentangle the personalized preference into a commonality part and a personality part from the network architecture perspective in this paper. The PFBP model is constructed with a universal feature extractor that represents aesthetic commonality and a personalized predictor that represents aesthetic personality. Specifically, the feature extractor is supervised by the average rated score which is similar to the training paradigm of a common FBP model, while the predictor is trained using individual rated score under a meta-learning paradigm. The predictor is expected a fast adaptation ability, but using conventional meta-learning paradigms are usually trapped in slow adaptation or over-fitting the tiny support set. To enhance its adaptation ability, we introduce learning-to-learn paradigm into a high-order predictor. Compared with the conventional predictor, aka first-order predictor, which is simply implemented by a fully-connected layer, the high-order predictor possesses a more powerful adaptation ability, by using a shallow parameter-generator to twist the weights of the predictor based on the input. Based on such architecture design, we further optimize the generator via a gradient-based meta-learning approach to form a meta-generator. Figure \ref{fig:method} illustrates the advantage of the proposed method, where the meta-generator can twist the weights of the high-order predictor quadratically for faster adaptation.

To stress the effectiveness of the proposed framework, termed \emph{MetaFBP}, we establish several PFBP benchmarks based on the existing FBP datasets varied from small, medium to large scales, including PFBP-SCUT500, PFBP-SCUT5500, and PFBP-US10K. We conduct extensive experiments on these newly-established PFBP benchmarks, and the experimental results demonstrate that our method significantly outperforms the conventional meta-learning approaches.  To summarize, the main contributions of this paper are concluded as follows:
\begin{enumerate}
\item[1)] Considering the nature of human aesthetic perception, we propose a disentangled training paradigm to study PFBP, which trains a \emph{universal feature extractor} to capture the aesthetic commonality and a \emph{personalized predictor} to adapt the aesthetic preferences of different users.
\item[2)] Based on the above training paradigm, we establish a \emph{MetaFBP} framework, which adopts a novel learning-to-learn mechanism to optimize the personalized predictor. Specifically, we introduce a high-order predictor and optimize a meta-generator to twist the weights of the predictor quadratically for fast adaptation.
\item[3)] We build several PFBP benchmarks based on the existing FBP datasets. Extensive experiments on these benchmarks demonstrate the effectiveness and superiority of the proposed method to conventional meta-learning approaches. Our method can act as a strong baseline to study PFBP in the future works. Both benchmark datasets and source code are available at: \url{https://github.com/MetaVisionLab/MetaFBP}.
\end{enumerate}

\section{Related Works}
\paragraph{\textbf{From Facial Beauty Prediction to Personalized Facial Beauty Prediction}}
The goal of FBP is to train a model as smart as humans to estimate facial attractiveness. Conventional approaches \cite{Aarabi,bottino2012intrinsic,PersonalizedFAP,kagian2007humanlike} tend to use geometric features or global appearance features (e.g., Color Histograms, Local Binary Pattern, Histogram of Oriented Gradients, Gabor Filters, etc.) to learn FBP.  However, such handcraft features heavily depend on heuristic rules.
Owing to the great success of deep learning \cite{abdel2014convolutional,ren2016faster,krizhevsky2017imagenet}, FBP can be easily optimized by Convolution Neural Network (CNNs) \cite{xie2015scut,fan2017label, liang2017region, lin2018r, lin2019regression} in an end-to-end manner. However, most methods for FBP are designed to learn population aesthetics. PFBP is much less explored. In order to prepare data for learning personalized facial attractiveness preferences, Whitehill et al. \cite{PersonalizedFAP} invited 8 volunteers to rate 1000 images. They trained regression models of facial beauty for each volunteer, and the experimental results indicated that personalized facial attractiveness preferences can be learnt by machine learning. Wang et al. \cite{wang2014demo} deemed that public aesthetic perception consisted of population aesthetics and personalized aesthetics. They decomposed the attractiveness score matrix into a low-rank matrix of population aesthetics and a matrix of personalized aesthetics, and used them to train regression models for learning population and personalized aesthetic jointly. Another study \cite{rothe2016some} focused on recommendation of personalized facial beauty for a large social website. Deep features of facial images extracted by a CNN are fed to collaborative filtering model. 
These works had validated that the subjective PFBP task can be solved by various machine learning methods. However, none of them study PFBP under a few-shot learning setting which is much more applicable in real-world scenarios.

\paragraph{\textbf{Few-Shot Learning}}
With the help of large-scale training data (\emph{e.g.}, ImageNet \cite{deng2009imagenet} and MS COCO \cite{lin2014microsoft}) and powerful computation resources, deep models have achieved great success \cite{radford2021learning,kirillov2023segment,chen2019all,chen2020unsupervised,chen2022label}. However, deep models may fail to rapidly generalize to new tasks when given a few examples. To tackle this challenge, meta-learning \cite{LearningtoLearn} is proposed as a new learning paradigm. The purpose of meta-learning is to learn to solve the unseen new task using meta knowledge from various tasks instead of singe task. Few-shot learning (FSL) \cite{wang2020generalizing}, as an application of meta-learning, can learn from a small number of examples even without them (zero-shot learning \cite{wang2018zero,narayan2020latent}). Researches of FSL have been greatly developed and can be categorized into many perspectives. Metric-based FSL \cite{vinyals2016matching,snell2017prototypical,sung2018learning,yoon2019tapnet,li2019finding} learns a representation space where similarities among samples are computed with a specific distance metric. Memory-based FSL \cite{santoro2016meta,munkhdalai2017meta,cai2018memory} stores the learned knowledge as key-value pairs by using a memory component where new samples are considered as a query to match the most similar key. Optimization-based FSL is to use prior knowledge to search parameters which generalize better to novel tasks \cite{nichol2018first,sun2019meta,kalais2022stochastic,yang2022efficient}. Finn et al. \cite{finn2017model} proposed a popular algorithm, MAML, to train the given neural network with a few gradient descent steps. To achieve this, MAML introduces two optimization loops for meta-learning, including an inner loop for task learning and an outer loop for training a meta-learner. The inner and outer loops are collaboratively optimized to find a meta-initialization that can be quickly adapted to different novel tasks. In this paper, we claim that PFBP is a more challenging task which requires a faster adaptation ability. To this end, for the first time, we upgrade the learning-to-learn mechanism with a high-order predictor and validate the significant superiority on PFBP.

\paragraph{\textbf{Personalized Image Aesthetics Assessment}} Personalized Image Aesthetics Assessment (PIAA) \cite{deng2017personalized,zhu2020personalized,zhu2021learning,li2020personality,yang2022personalized} aims to learn to assess the aesthetic quality (or score) of images by taking into account the users' aesthetic preferences. PIAA is a recent popular topic which is derived from the Generic Image Aesthetic Assessment (GIAA) \cite{zhang2019gated,cui2018distribution}. The PIAA are more related to our work as it learns personalized aesthetics for the image quality.
Most PIAA works attempt to learn the individual aesthetic assessment by exploiting and transferring the learned knowledge from trained GIAA model \cite{deng2017personalized, li2020personality}, or using extra supervision information \cite{zhu2021learning,li2019personality}. In this kind of task, the personalized aesthetics models are optimized to quickly adapt to a new user's aesthetic preference, and these PIAA models may fail to capture personalized aesthetics \cite{zhu2020personalized}. To this end, recent PIAA works \cite{wang2019meta, zhu2020personalized,li2022transductive,yang2022personalized} based on meta-learning paradigms are proposed to tackle this problem. Although the promising PIAA performance is achieved, most methods still have complex training frameworks \cite{li2019personality,wang2019meta,zhu2020personalized} that are not suitable for deployment in practice. Furthermore, existing methods mainly focus on FSL tasks with larger shots (10-shot and 100-shot), which means more labeled images are necessary for model fine-tuning. In this paper, we explore learning the personalized aesthetics for facial attractiveness with less supervision information in the standard FSL setting, leading to an urgent requirement on the fast adaptability using extremely-limited labeled examples.

\begin{figure*}[ht]
\centering
\includegraphics[width=0.925\textwidth]{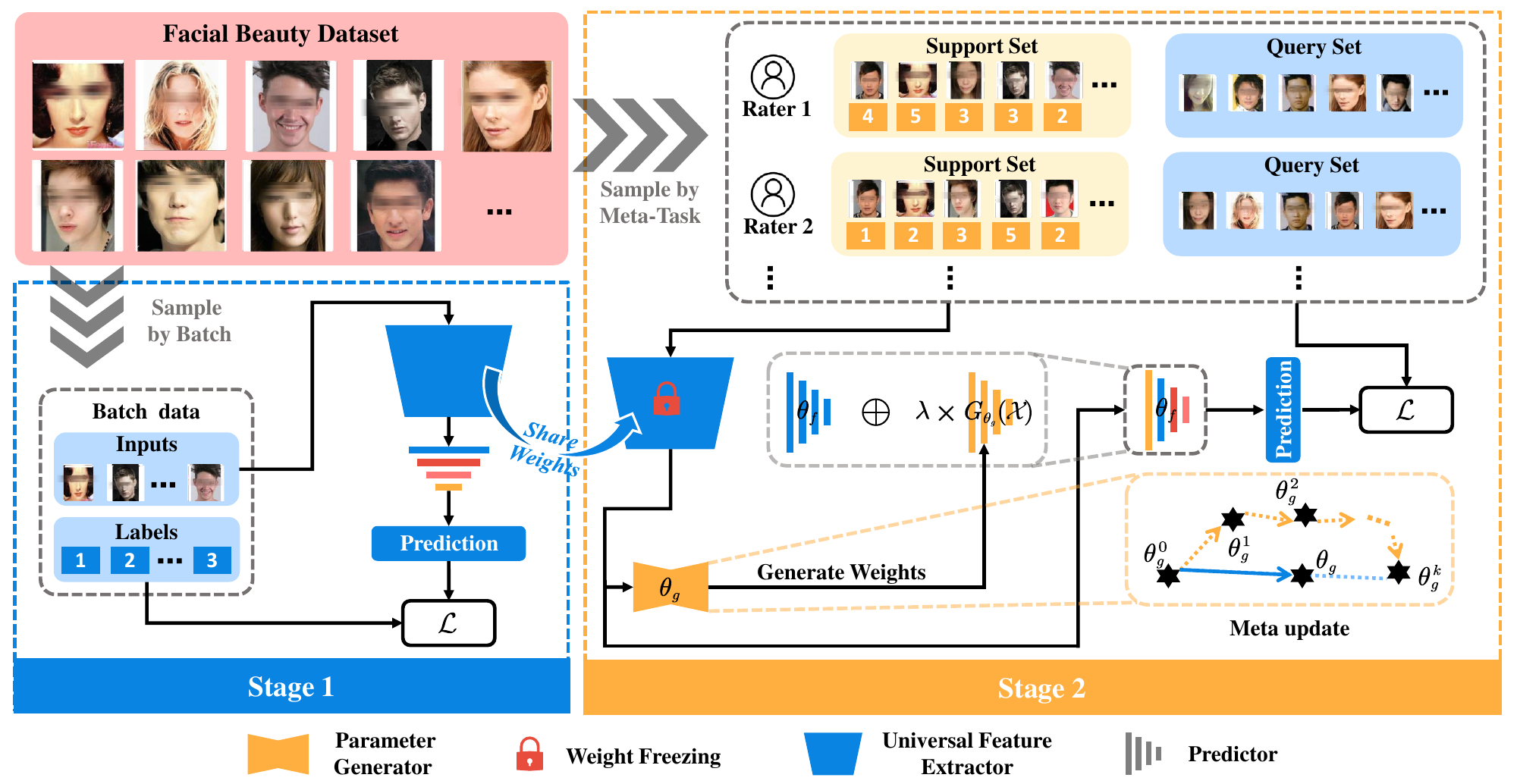}
\vskip -0.1in
\caption{The illustration of the proposed MetaFBP framework. 1) Stage 1: Train a universal feature extractor. 2) Stage 2: Train a personalized high-order predictor, which is designed with a parameter generator $\theta_g$. We further optimize $\theta_g$ via meta-learning to form a meta-generator so as to adapt to different new users given limited labeled images.}
\vskip -0.1in
\label{fig:pipeline}
\end{figure*}

\section{Task Formulation}
As mentioned above, a PFBP system can be arranged in the following manner: the system requires each user to label a few facial images that demonstrate their aesthetic preferences. The model is then fine-tuned on such limited labeled data for task adaptation. To meet this application manner, PFBP is modeled as a meta-learning paradigm in this paper. Specifically, each user corresponds to a meta-task. And all the meta-tasks are randomly divided into meta-train tasks and meta-test tasks. In each meta-task, the images are divided into a support set and a query set. In the meta-train tasks, the query set is a pseudo query set for meta-learning. In the meta-test tasks, the query set is used for performance evaluation without annotations. To provide sufficient aesthetic information in the support set for training, we urge each user to select the images from the gallery to traverse $C$ different attractiveness scores with $K$ samples per score, termed as ``C-way K-shot''. 

\paragraph{\textbf{Notations}}
Figure \ref{fig:intro_fig1} illustrates an example that formulates PFBP as meta-tasks. We represent the meta-train set as $\mathcal{D}_{train}=\{\mathcal{D}_m\}_{m=1}^M$, where $M$ denotes the number of users and $\mathcal{D}_m$ denotes the data rated by the $m$-th user. Then, a meta-task $\mathcal{T}_m$ is sampled from $\mathcal{D}_m=\{(\mathcal{X}_m, \mathcal{Y}_m)\}$ that contains the images $\mathcal{X}_m$ and the corresponding beauty scores $\mathcal{Y}_m$ rated by the $m$-th user. Each meta-task consists of a support set $\mathcal{S}_m=\{(\mathcal{X}_m^s, \mathcal{Y}_m^s)\}$ and a query set $\mathcal{Q}_m=\{(\mathcal{X}_m^q, \mathcal{Y}_m^q)\}$, where $\mathcal{X}_m^s, \mathcal{X}_m^q \in \mathcal{X}_m$ and $\mathcal{Y}_m^s, \mathcal{Y}_m^q \in \mathcal{Y}_m$. The support set $\mathcal{S}_m$ and query set $\mathcal{Q}_m$ are constructed by randomly selecting $N_s$ and $N_q$ samples from the subset $\mathcal{D}_m$ for each attractiveness score without overlapping. Similarly, the meta-test set is built in the same way, where the support set is used for model fine-tuning, but leaving the unlabeled query set to evaluate the adaptation performance of the model fine-tuned by the support set. 

\paragraph{\textbf{Training objective}}
Since our method is strongly correlated with the optimization-based meta-learning methods~\cite{finn2017model}, we only present the training pipeline of the optimization-based methods for the PFBP task. For each meta-task/episode, we first update a model $F(\cdot)$ on the support set $\mathcal{S}_m$ by calculating the regression loss, aka the mean squared error (MSE) loss $\mathcal{L}_{mse}(F(\mathcal{X}_m^s), \mathcal{Y}_m^s)$,  and then use the updated model to predict the beauty scores of the query images $\mathcal{X}_m^q$. The predicted scores can be formulated as follows:
\begin{equation}
    \widehat{\mathcal{Y}}_m^q = F(\mathcal{X}_m^q |\nabla \mathcal{L}_{mse}(F(\mathcal{X}_m^s), \mathcal{Y}_m^s)).
\end{equation}
Subsequently, the training objective of each meta-task is to minimize the regression loss between the predicted scores and the corresponding ground-truth rating labels:
\begin{equation}
    \min \limits_F \mathcal{L}_{mse}(\widehat{\mathcal{Y}}_m^q, \mathcal{Y}_m^q).
    \label{eq:loss_on_query}
\end{equation}

\section{Method}
The human aesthetic preference can be disentangled into a commonality part and a personality part. The former represents the consistent judgement of the majority while the latter represents the individual variations from the majority. To meet this prior knowledge, we decompose the network architecture into a universal feature extractor to capture the aesthetic commonality and a predictor to capture the aesthetic personality by shifting the decision boundary for task adaptation. To enhance task adaptation, we introduce a high-order predictor which updates the predictor by using a shallow parameter-generator network. We further optimize the generator into a meta-generator via meta-learning.  These two components (commonality \emph{vs.} personality) are optimized independently with different optimization objectives. The whole training process is illustrated in Figure \ref{fig:pipeline}.

\subsection{Stage 1: Universal Feature Extractor}
In the first stage, we aim to learn a universal feature extractor to capture aesthetic commonality by training a common FBP model, which is composed of a feature extractor $E_{\theta_e}$ and a predictor $F_{\theta_f}$. This model is trained identical to the common FBP training manner. Since the labels used in this stage are required to represent aesthetic commonality, we choose \emph{mode} rating to ensemble the scores from all users, that is, the most-popular score in the rating distribution is exploited as the true label of each image. Take an image $x$ as an example, its rating distribution is represented as $\{y_1, ..., y_m, ...,y_M\}$ where $y_m$ is the attractiveness score rated by the $m$-th user. The label representing the aesthetic commonality is defined as:
\begin{equation}
    \overline{y} = \arg \max \limits_{c} \sum \limits_{m=1}^M \delta(y_m=c),
\end{equation}
where $c$ ranges from 1 to $C$. \emph{mode} rating is considered more representative than \emph{mean} rating to capture aesthetic commonality, since it does not take into account the opinions of minorities. Using $\overline{y}$ as the supervision signal, the training goal can be formulated as:
\begin{equation}
    \min\limits_{\theta_e, \theta_f} \sum_{(x,\overline{y})\in\mathcal{D}_{train}} \mathcal{L}(F_{\theta_f}(E_{\theta_e}(x)), \overline{y}),
\end{equation}
where $\mathcal{L}$ is a prediction loss. Upon completion of the training stage for the common FBP model, the predictor $F_{\theta_f}$ is discarded, and only the universal feature extractor $E_{\theta_e}$ is retained. It is worth noting that in the subsequent stage, the weights of the universal feature extractor are fixed, with the goal of maintaining the knowledge of aesthetic commonality across different meta-tasks.

\subsection{Stage 2: Personalized High-Order Predictor}
Based on the commonality-aware feature extractor in the first stage, we need a personalized predictor to meet various user preferences by shifting the decision boundary adaptively. One straightforward approach is to use the conventional meta-learning methods like MAML~\cite{finn2017model} to optimize the predictor. However, PFBP is more challenging than previous image recognition tasks, which requires a faster adaptation ability, while the conventional meta-learning methods usually result in slow adaptation or overfitting the tiny support sets. As an alternative, we propose a high-order predictor that is dynamically twisted conditioned on the inputs, by using a shallow parameter-generator. We optimize the generator via a gradient-based meta-learning method to achieve a meta-generator which can adapt the meta-tasks more quickly. See Figure \ref{fig:method} to compare with conventional meta-learning mechanism. The implementation details are shown in Figure \ref{fig:pipeline} and Algorithm \ref{algm:MAMP}.

\paragraph{\textbf{High-Order Predictor}} In this paper, the predictor is implemented as a fully-connected (FC) layer with weights $\theta_f$. To enhance adaptation ability, a high-order predictor is encouraged here. Specifically, the weights $\theta_f$ in the high-order predictor can be twisted adaptively in test-time, which is formulated as:
\begin{equation}
    \theta_f = \theta_f + \lambda G_{\theta_g}(\mathcal{X}),
    \label{eq:mix_parameres}
\end{equation}
where $G$ is a shallow parameter generator conditioned on the input features $\mathcal{X}$ provided by the feature extractor. $\theta_g$ is the weight of the generator. $\lambda$ is a hyper-parameter that represents the adaptation strength. In practice, the parameter generator is implemented as a multi-layer-perceptron with the structure of FC-ReLU-FC. Compared with the first-order predictor, the high-order predictor has a much higher freedom for task adaptation.

\paragraph{\textbf{Meta-Generator}} Based on the design of high-order predictor, we aim to further optimize the parameter-generator into a meta-generator via meta-learning. Roughly speaking, in the inner loop, we conduct $k$-step adaptation to optimize $\theta_g$ on the support set $\mathcal{S}_m$ to achieve a ghosted $\theta_g$, termed $\theta_g'$. In the outer loop, we conduct meta-update to $\theta_g$ on the query set $\mathcal{Q}_m$ by using $\theta_g'$. In detail, the prediction of the high-order predictor on the support set can be represented as $F_{\theta_f \circ \theta_g}(\mathcal{X}_m^s)$. The weights $\theta_g$ are updated based on the gradients of the model on the support set $\mathcal{S}_m$:
\begin{equation}
    \theta_g' \leftarrow \theta_g - \alpha \underbrace{\nabla_{\theta_g} \mathcal{L} (F_{\theta_f \circ \theta_g}(\mathcal{X}_m^s), \mathcal{Y}_m^s)}_{k-step},
    \label{eq:maml_1}
\end{equation}
where $\alpha$ is a hyper-parameter representing the step size of inner update. The process in Equation \ref{eq:maml_1} is repeated $k$ times to obtain a more task-oriented gradient that makes the adaptation more thoroughly. We then calculate the quadratic gradients of the updated predictor on the query set to update $\theta_g$ by taking $\theta_g'$ as a bridge:
\begin{equation}
    [\theta_f, \theta_g] \leftarrow [\theta_f, \theta_g] - \beta \nabla_{\theta_f, \theta_g} \mathcal{L} (F_{\theta_f \circ \theta_g'}(\mathcal{X}_m^q), \mathcal{Y}_m^q),
    \label{eq:maml_2}
\end{equation}
where $\beta$ is a hyper-parameter denoting the step size of outer loop. Note that we update the generator based on its initial weights to obtain a generalizable initial weights. Different from $\theta_g$, $\theta_f$ is optimized in a standard training manner.

\begin{algorithm}[!t]
    \SetAlgoLined
	\caption{Learning-to-learn high-order predictor} \label{algm:MAMP}
    \algorithmicrequire{Training data $\mathcal{D}_{train}=\{\mathcal{D}_m\}_{m=1}^M$, $k$-step in the inner loop, training iteration $I$} \\
    Randomly initialize weights $\theta_f$, $\theta_g$ \;
    \For{$i \leftarrow 1$ \KwTo ${I}$}{
        Sample a meta-task $\mathcal{T}_m$ from $D_m$, $m\in [1,M]$\;
            Acquire the support set $\mathcal{S}_m \sim \mathcal{T}_m$ \;
            $\theta_g' \leftarrow \theta_g$ \;
            \Repeat {$k$ times} {
                Update $\theta_g'$ on $\mathcal{S}_m$ using Equation \ref{eq:maml_1} \;
            }
            Acquire the query set $\mathcal{Q}_m \sim \mathcal{T}_m$ \;
            Update $\theta_f, \theta_g$ on $\mathcal{Q}_m$ using Equation \ref{eq:maml_2} \;
        }
\end{algorithm}

\section{Experiments}
\subsection{Experimental Benchmarks}
We construct three PFBP benchmarks varying from small to large scales, based on the public FBP datasets that provide beauty scores annotated by multiple raters. 
(1) \textbf{PFBP-SCUT5500} is collected from SCUT-FBP5500~\cite{liang2017SCUT}, which consists of 5,500 face images, each labeled by 60 users. The images vary widely in terms of characteristics such as gender and ethnicity, making it difficult to predict esthetic preferences. (2) \textbf{PFBP-SCUT500} is collected from SCUT-FBP500 \cite{xie2015scut}, which contains 500 facial images from the Asian female population rated by 75 volunteers. (3) \textbf{PFBP-US10K} is sampled from 10K US Adult Faces dataset~\cite{bottino2012intrinsic}, consisting 2,222 facial images from Caucasian population rated by 12 users. All the above datasets are annotated at a beauty scale of $\{1,2,3,4,5\}$, where the higher beauty score represents the more attractive face.

\paragraph{\textbf{Dataset Split}}
\begin{table}[t]
  \centering
  \caption{The details of dataset split.}
  \label{tab:dataset_statistics}%
   \vspace{-10pt}
  \setlength\tabcolsep{9pt}\scalebox{0.80}{
    \begin{tabular}{c|cccc}
    \toprule[1pt]
    \multirow{2}[2]{*}{Item} & \multicolumn{4}{c}{PFBP-SCUT5500} \\
          & Total & Train Set & Validation Set & Test Set \\
    \midrule
    Number of images & 500   & 300   & 50    & 150 \\
    Number of users & 60    & 30    & 10    & 20 \\
    Total annotations & 12,500 & 9,000 & 500   & 3,000 \\
    \midrule[1pt]
    \multirow{2}[2]{*}{Item} & \multicolumn{4}{c}{PFBP-SCUT500} \\
          & Total & Train Set & Validation Set & Test Set \\
    \midrule
    Number of images & 5,500  & 3,000  & 500   & 2,000 \\
    Number of users & 60    & 30    & 10    & 20 \\
    Total annotations & 135,000 & 90,000 & 5,000 & 40,000 \\
    \midrule[1pt]
    \multirow{2}[2]{*}{Item} & \multicolumn{4}{c}{PFBP-US10K} \\
          & Total & Train Set & Validation Set & Test Set \\
    \midrule
    Number of images & 2,222  & 1,111  & 667   & 444 \\
    Number of users & 12    & 6     & 2     & 4 \\
    Total annotations & 9,771 & 6,663  & 1,333  & 1,775 \\
    \bottomrule[1pt]
    \end{tabular}%
    }
    \vspace{-0.2cm}
\end{table}%
We split each dataset into train, validation and test sets in a 6:3:1 ratio based on the users who provided the annotations. Additionally, we ensure that the images in each split are distinct. Note that some users provided empty ratings for the extremely high or low beauty score, making it difficult to sample meta-tasks from these users. As a result, we exclude such users from the split. The details of dataset split are listed in Table \ref{tab:dataset_statistics}. 

\paragraph{\textbf{Evaluation Protocols}}
The model that performed the best on the validation set is chosen for evaluation. We use Pearson correlation (PC), mean absolute error (MAE), and root mean squared error (RMSE) to measure the regression performance of our method. A higher PC, smaller MAE and RMSE indicate the better performance achieved by the model on PFBP task.

\subsection{Experimental Details}
\paragraph{\textbf{Setup}}
We follow the meta-training and meta-testing settings from few-shot learning tasks to conduct experiments. Specifically, we perform 5-way K-shot regression meta-tasks on PFBP-SCUT5500 and PFBP-US10K. However, due to a large number of empty annotations for certain categories in PFBP-SCUT500, we rearrange the score labels by reducing the beauty category number from 5 to 3 via score mapping: $\{1,2\} \rightarrow 1, \{3\} \rightarrow 2, \{4,5\} \rightarrow 3$. Afterwards, we conduct 3-way K-shot regression meta-tasks on the PFBP-SCUT500.

\paragraph{\textbf{Cyclically Re-sampling Strategy}}
When performing each meta-task, we need to select $N_s + N_q$ images per category to create the support set and the query set. However, user ratings are usually imbalanced, with a few samples receiving scores of 1 and 5, while thousands receive a score of 2. This can lead to an extreme situation where the number of images in the minority categories is insufficient to create a meta-task. To solve this problem, we devise a Cyclically Re-sampling Strategy for the minority categories. Assuming $N_c$ denotes the sample number of the $c$-th category rated by the $m$-th user, the sampling strategy is defined as follows:

\textbf{Case 1}: If $N_c=1$,  we duplicate the single sample $N_s$ times to form the $c$-th category in the support set, and the $c$-th category in the query set will be left empty.

\textbf{Case 2}: If $1<N_c\leq N_s$, we first randomly select $N_c-1$ samples from the $c$-th category to form the support set by duplicated sampling $K/(N_c-1)$ times, and the remaining one is duplicated sampling $N_q$ times to form the query set.

\textbf{Case 3}: If $N_s<N_c\leq N_s+N_q$, we first randomly select $N_s$ samples from the $c$-th category to form the support set, and the remaining samples are duplicated sampling $N_q/(N_c-N_s)$ times to form the query set.

\textbf{Case 4}: If $N_c> N_s+N_q$, we first randomly select $N_s$ samples from the $c$-th category to form the support set, and then randomly select $N_q$ samples from the remaining ones to form the query set.

\begin{table*}[t]
  \centering
 \caption{5-way K-shot regression results on PFBP-SCUT5500 benchmark.
  The same number of shots is kept during both meta-training and meta-testing phases.
  The best and second-best results are marked by \textbf{bold} and \underline{underline}, respectively. Same representation in the following tables.}
  \label{tab:fbp5500}%
   \vspace{-10pt}
  \setlength\tabcolsep{13pt}\scalebox{0.80}{
    \begin{tabular}{c|c|ccc|ccc|ccc}
    \toprule
    \multirow{2}[2]{*}{Type} & \multirow{2}[2]{*}{Method} & \multicolumn{3}{c|}{1 Shot} & \multicolumn{3}{c|}{5 Shot} & \multicolumn{3}{c}{10 Shot} \\
          &       & PC    & MAE   & RMSE  & PC    & MAE   & RMSE  & PC    & MAE   & RMSE \\
    \midrule
    \multirow{2}[2]{*}{Baseline} & Base-CommonFBP & 0.6827 & 0.8668 & 1.1135 & 0.7812 & 0.7088 & 0.9044 & 0.7992 & 0.6826 & 0.8481 \\
          & Base-MAML \cite{finn2017model} & 0.7549 & 0.8480 & 1.1268 & 0.7837 & 0.7766 & 1.0245 & 0.7862 & 0.7906 & 1.0249 \\
    \midrule
    \multirow{2}[2]{*}{FSL} & ProtoNet \cite{snell2017prototypical} & 0.7816 & 0.7053 & 0.8620 & 0.7969 & 0.6838 & 0.8450 & 0.7980 & 0.6932 & 0.8875 \\
          & MTL \cite{sun2019meta} & 0.7228 & 0.8593 & 1.0972 & 0.7350 & 0.9040 & 1.1832 & 0.7277 & 0.8924 & 1.1375 \\
    \midrule
    PIAA  & BLG-PIAA \cite{zhu2020personalized} & 0.7927 & 0.6850 & 0.8426 & 0.7683 & 0.7853 & 1.1154 & 0.7705 & 0.7995 & 1.0434 \\
    \midrule
    \multirow{2}[2]{*}{Ours} & MetaFBP-R & \underline{0.8037} & \underline{0.6780} & \underline{0.8365} & \underline{0.8050} & \underline{0.6727} & \underline{0.8318} & \underline{0.8098} & \underline{0.6631} & \underline{0.8208} \\
          & MetaFBP-T & \textbf{0.8067} & \textbf{0.6701} & \textbf{0.8274} & \textbf{0.8061} & \textbf{0.6716} & \textbf{0.8282} & \textbf{0.8125} & \textbf{0.6572} & \textbf{0.8147} \\
    \bottomrule
    \end{tabular}%
  }
  \vskip -0.1in
\end{table*}%
\begin{table*}[t]
  \centering
  \caption{3-way K-shot regression results on PFBP-SCUT500 benchmark.}
  \label{tab:fbpscut}%
   \vspace{-10pt}
  \setlength\tabcolsep{13pt}\scalebox{0.80}{
    \begin{tabular}{c|c|ccc|ccc|ccc}
    \toprule
    \multirow{2}[2]{*}{Type} & \multirow{2}[2]{*}{Method} & \multicolumn{3}{c|}{1 Shot} & \multicolumn{3}{c|}{5 Shot} & \multicolumn{3}{c}{10 Shot} \\
          &       & PC    & MAE   & RMSE  & PC    & MAE   & RMSE  & PC    & MAE   & RMSE \\
    \midrule
    \multirow{2}[2]{*}{Baseline} & Base-CommonFBP & 0.5324 & 0.5335 & 0.6658 & 0.7206 & 0.4379 & 0.5491 & 0.7525 & 0.4444 & 0.5342 \\
          & Base-MAML \cite{finn2017model} & 0.7074 & 0.4244 & 0.5659 & 0.7688 & 0.3915 & 0.5040 & 0.7708 & 0.3840 & 0.4996 \\
    \midrule
    \multirow{2}[2]{*}{FSL} & ProtoNet \cite{snell2017prototypical} & 0.6269 & 0.5279 & 0.5920 & 0.7693 & 0.3907 & 0.5081 & 0.7694 & 0.3865 & 0.5175 \\
          & MTL \cite{sun2019meta} & 0.6706 & 0.5196 & 0.7156 & 0.6739 & 0.5192 & 0.7152 & 0.6645 & 0.5157 & 0.7121 \\
    \midrule
    PIAA  & BLG-PIAA \cite{zhu2020personalized} & 0.7139 & 0.3995 & 0.5439 & 0.7369 & 0.4029 & 0.5372 & 0.7577 & 0.3995 & 0.5238 \\
    \midrule
    \multirow{2}[2]{*}{Ours} & MetaFBP-R & \textbf{0.7478} & \textbf{0.3840} & \textbf{0.5335} & \textbf{0.7769} & \textbf{0.3720} & \underline{0.5026} & \underline{0.7729} & \underline{0.3772} & \underline{0.4960} \\
          & MetaFBP-T & \underline{0.7393} & \underline{0.3947} & \underline{0.5378} & \underline{0.7738} & \underline{0.3775} & \textbf{0.5025} & \textbf{0.7787} & \textbf{0.3746} & \textbf{0.4911} \\
    \bottomrule
    \end{tabular}%
    }
    \vskip -0.1in
\end{table*}%
\begin{table*}[!t]
  \centering
  \caption{5-way K-shot regression results on PFBP-US10K benchmark.}
  \label{tab:us10k}%
   \vspace{-10pt}
  \setlength\tabcolsep{13pt}\scalebox{0.80}{
    \begin{tabular}{c|c|ccc|ccc|ccc}
    \toprule
    \multirow{2}[2]{*}{Type} & \multirow{2}[2]{*}{Method} & \multicolumn{3}{c|}{1 Shot} & \multicolumn{3}{c|}{5 Shot} & \multicolumn{3}{c}{10 Shot} \\
          &       & PC    & MAE   & RMSE  & PC    & MAE   & RMSE  & PC    & MAE   & RMSE \\
    \midrule
    \multirow{2}[2]{*}{Baseline} & Base-CommonFBP & 0.2640 & 1.2355 & 1.5295 & 0.3871 & 1.1094 & 1.3507 & 0.4742 & 1.0473 & 1.2491 \\
          & Base-MAML \cite{finn2017model} & 0.4028 & 1.2719 & 1.6584 & 0.4286 & 1.2425 & 1.5975 & 0.4570 & 1.2324 & 1.5621 \\
    \midrule
    \multirow{2}[2]{*}{FSL} & ProtoNet \cite{snell2017prototypical} & 0.3094 & 1.2147 & 1.5943 & 0.4877 & 1.0260 & 1.2683 & 0.4906 & 1.0244 & 1.2389 \\
          & MTL \cite{sun2019meta} & 0.4678 & 1.0311 & 1.3136 & 0.4642 & 1.0322 & 1.3137 & 0.4610 & 1.0594 & 1.3774 \\
    \midrule
    PIAA  & BLG-PIAA \cite{zhu2020personalized} & 0.4691 & 1.0281 & 1.3066 & 0.4615 & 1.0476 & 1.3599 & 0.4731 & 1.0502 & 1.3339 \\
    \midrule
    \multirow{2}[2]{*}{Ours} & MetaFBP-R & \underline{0.5067} & \underline{0.9970} & \underline{1.2356} & \underline{0.4968} & \underline{1.0121} & \textbf{1.2344} & \textbf{0.5004} & \textbf{1.0115} & \textbf{1.2290} \\
          & MetaFBP-T & \textbf{0.5109} & \textbf{0.9968} & \textbf{1.2182} & \textbf{0.5007} & \textbf{1.0086} & \underline{1.2425} & \underline{0.4991} & \underline{1.0128} & \underline{1.2322} \\
    \bottomrule
    \end{tabular}%
    }
  \vskip -0.1in
\end{table*}%

\paragraph{\textbf{Implementation Details}}
Our experiments are implemented on Pytorch platform and runs on a NVIDIA RTX3090 GPU. For all experiments, we use ResNet-18~\cite{he2015deep} as the network backbone, which is initialized by the ImageNet pre-trained model~\cite{deng2009imagenet}. Before training, we simply apply several augmentation techniques to preprocess images, including random crop and random horizontal flipping. During the first training stage, we train the universal feature extractor by using cross-entropy loss and SGD optimizer with batchsize of 64, maximum epochs of 100, and a learning rate of 0.001 stepped down by half per 20 epochs.
In the second stage, we freeze the weights of the universal feature extractor, and develop a high-order predictor by implementing a meta-generator that is a MLP with structure of FC-ReLU-FC. The hyper-parameter of adaptation strength $\lambda$ is set to 0.01. The inner-update step size $\alpha$, outer-update step size $\beta$ and the number of $k$-step are set to 0.01, 0.001 and 10, respectively. And we sample 40,000 and 400 meta-tasks from the training set and the testing set for meta-training and meta-testing, respectively. Note that the shot number of the support set is kept consistent in both the meta-training and meta-testing phases, if without additional explanation. And the shot number of the query set is set to 15 by default.

\paragraph{\textbf{Strong Baselines}}
As the first time applying meta-learning to formulate PFBP, there are currently no experimental results on these new benchmarks. To demonstrate the effectiveness of our proposed method, we also implement several strong baselines on these PFBP benchmarks, including:
1) \textbf{Base-commonFBP}:
In line with conventional training methods~\cite{liang2017SCUT}, we developed a common FBP model with the same architecture as our model. The common FBP model can represent the aesthetic commanlity, which is assumed to be correlated with user preferences to some extent. We then evaluate its effectiveness on the PFBP task using the same meta-testing manner as our approach.
2) \textbf{Base-MAML}:
MAML~\cite{finn2017model} is a popular meta-learning approach to address few-shot learning tasks. To highlight the advantage of our method, we also implement MAML on the PFBP task. For a fair comparison, MAML is implemented using the same task formulation, architecture, and hyper-parameter settings (e.g., $\alpha, \beta, \gamma$) as our method. 

\paragraph{\textbf{Other Related Methods}}
In order to provide a comprehensive evaluation of the proposed method, we also compare it with other state-of-the-art methods on our PFBP task. Specifically, we re-implement \textbf{FSL} methods, i.e., ProtoNet \cite{snell2017prototypical} and MTL \cite{sun2019meta}, and a recent \textbf{PIAA} method, i.e., BLG-PIAA \cite{zhu2020personalized} on PFBP task. Since ProtoNet is originally designed for few-shot classification tasks, we modify it for PFBP task by calculating the expectation score of the output distribution as the final prediction result.

\subsection{Experimental Results}
To further investigate the impact of different parameter updating manners on the high-order predictor, we implement two different methods, known as parameter-tuning and parameter-rebirth. 
Parameter tuning aims to modulate the parameters of the predictor by generating dynamic residuals to add to the original parameters. The operation of parameter tuning is illustrated in Equation \ref{eq:mix_parameres}.
Unlike parameter tuning, parameter rebirth discards the original parameters and generates the new parameters by the parameter generator $G_{\theta_g}$ conditioned on the input features $\mathcal{X}$. The operation of parameter rebirth can be formulated as: $\theta_f = G_{\theta_g}(\mathcal{X})$.
For simplicity, our method implemented with parameter tuning and parameter rebirth are termed as \textbf{MetaFBP-T} and \textbf{MetaFBP-R}, respectively. 
\paragraph{\textbf{Comparison with Strong Baselines}}
Our method is compared with the strong baselines (\emph{i.e.}, Base-commonFBP and Base-MAML) to stress its effectiveness on PFBP. The comparison results on PFBP-SCUT5500, PFBP-SCUT500 and PFBP-US10K benchmarks in terms of PC, MAE and RMSE are reported in Table \ref{tab:fbp5500}, Table \ref{tab:fbpscut} and Table \ref{tab:us10k}, respectively. From these tables, we can observe that our method almost surpasses all the strong baselines with a much higher PC and smaller MSE, RMSE over all the benchmarks, in terms of different K-shot settings. For the most challenging 1-shot setting, our method (MetaFBP-R and MetaFBP-T) both achieve a great PC improvements of more than 4\% and 3\% on the PFBP-SCUT5500 and PFBP-SCUT500 benchmarks, respectively, compared to the baselines.
Moreover, our method demonstrates a more significant performance improvement over Base-MAML on the user-less PFBP-US10K benchmark, with an improvement in PC more than 10\%. This result highlights the ability of our method to adapt to new tasks even when training data is limited. Furthermore, as the number of training shots increases, the performance of our method improves correspondingly, with the most significant improvement observed in the 1-shot setting. In practically, the improvement in 1-shot setting is particularly significant in real-world scenarios as it allows for a more convenient user experience with fewer required ratings.
\vspace{-0.3cm}
\paragraph{\textbf{Comparison with Other Related Methods}}
From Table \ref{tab:fbp5500}-\ref{tab:us10k}, we can observe that our method achieves the state-of-the-art results even compared with competitive methods, including PIAA and FSL methods, among all benchmarks, which demonstrates the effectiveness of our metaFBP method on PFBP task.

\begin{table}[t]
  \centering
  \caption{Ablation study of different K-shot settings during training and testing phases on PFBP-SCUT5500 dataset.}
    \vspace{-0.4cm}
  \label{tab:fbp5500_diffkshot}%
    \setlength\tabcolsep{12pt}\scalebox{0.80}{
    \begin{tabular}{c|c|ccc}
    \toprule
    \multirow{2}[2]{*}{Method} & \multirow{2}[2]{*}{Training Phase} & \multicolumn{3}{c}{Testing Phase} \\
          &       & 1 shot & 5 shot & 10 shot \\
    \midrule
    Base-MAML  & \multirow{2}[2]{*}{1 shot} & 0.7549 & 0.7690 & 0.7809 \\
    MetaFBP-T &       & \textbf{0.8067} & \textbf{0.8050} & \textbf{0.8104} \\
    \midrule
     Base-MAML  & \multirow{2}[2]{*}{5 shot} & 0.7495 & 0.7837 & 0.7898 \\
     MetaFBP-T &       & \textbf{0.8033} & \textbf{0.8061} & \textbf{0.8118} \\
    \midrule
    Base-MAML  & \multirow{2}[2]{*}{10 shot} & 0.7309 & 0.7758 & 0.7862 \\
    MetaFBP-T &       & \textbf{0.7914} & \textbf{0.8057} & \textbf{0.8125} \\
    \bottomrule
    \end{tabular}%
    }
    \vskip -0.15in
\end{table}%
\begin{table}[t]
  \centering
  \caption{Ablation study of the adaptation strength $\lambda$.}
    \vspace{-0.4cm}
  \label{tab:ablation_alpha}%
    \setlength\tabcolsep{25pt}\scalebox{0.80}{
    \begin{tabular}{c|cc|c}
    \toprule
    $\lambda$ & 1 shot & 5 shot & Avg \\
    \midrule
    1     & 0.5915 & 0.2026 & 0.3971 \\
    0.1   & 0.8059 & 0.7231 & 0.7645 \\
    0.01  & 0.\textbf{8067} & \textbf{0.8061} & \textbf{0.8064} \\
    0.001 & 0.6846 & 0.7575 & 0.7211 \\
    0.0001 & 0.4932 & 0.6932 & 0.5932 \\
    \bottomrule
    \end{tabular}%
    }
    \vskip -0.1in
\end{table}%
\begin{figure}[t]
     \centering
     \begin{subfigure}[b]{0.48\columnwidth}
         \centering
         \includegraphics[width=\columnwidth]{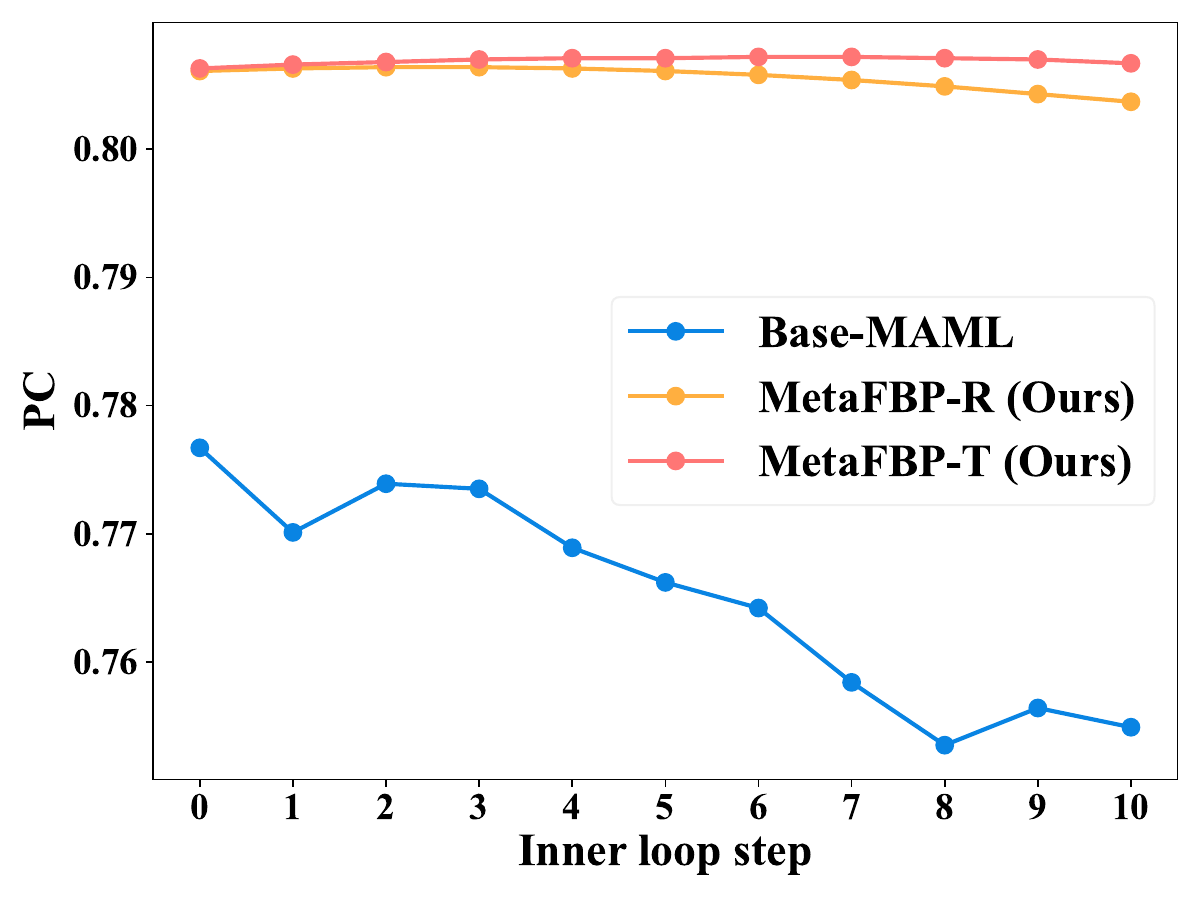}
         \vspace{-0.5cm}
         \caption{Train/Test with 1-shot.}
         \label{fig:mete_update_1shot}
     \end{subfigure}
     \begin{subfigure}[b]{0.48\columnwidth}
         \centering
         \includegraphics[width=\columnwidth]{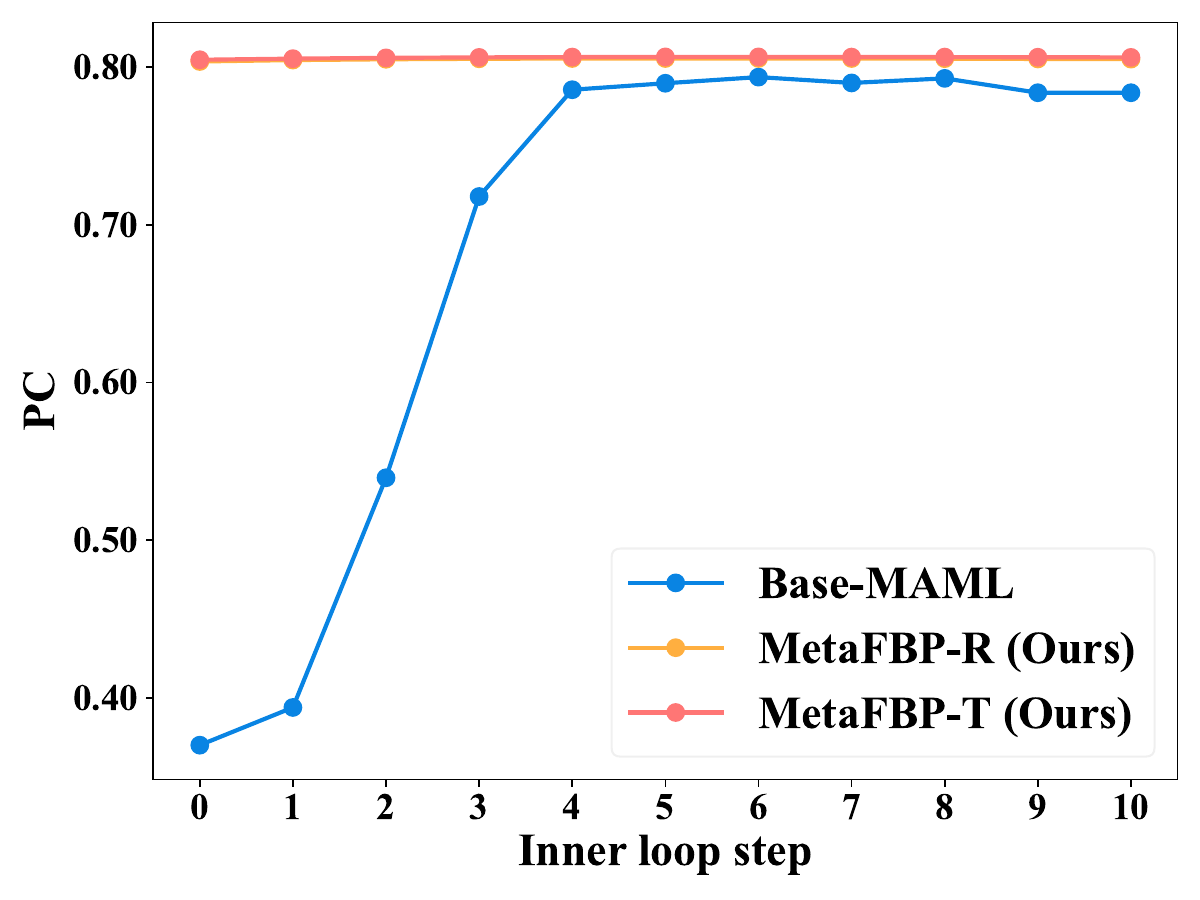}
         \vspace{-0.5cm}
         \caption{Train/Test with 5-shot.}
         \label{fig:mete_update_5shot}
     \end{subfigure}
    \begin{subfigure}[b]{0.48\columnwidth}
         \centering
         \includegraphics[width=\columnwidth]{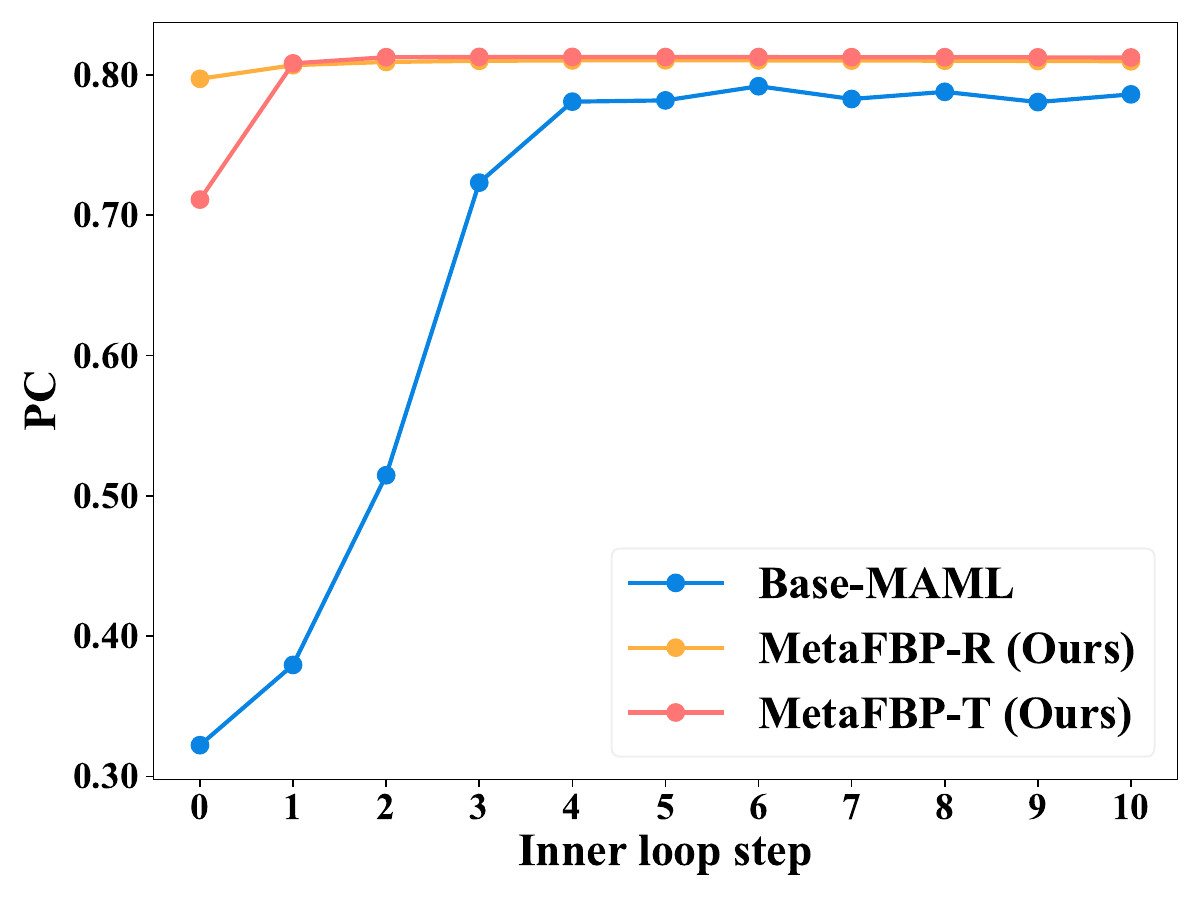}
         \vspace{-0.5cm}
         \caption{Train/Test with 10-shot.}
         \label{fig:mete_update_10shot}
    \end{subfigure}
    \begin{subfigure}[b]{0.48\columnwidth}
         \centering
         \includegraphics[width=\columnwidth]{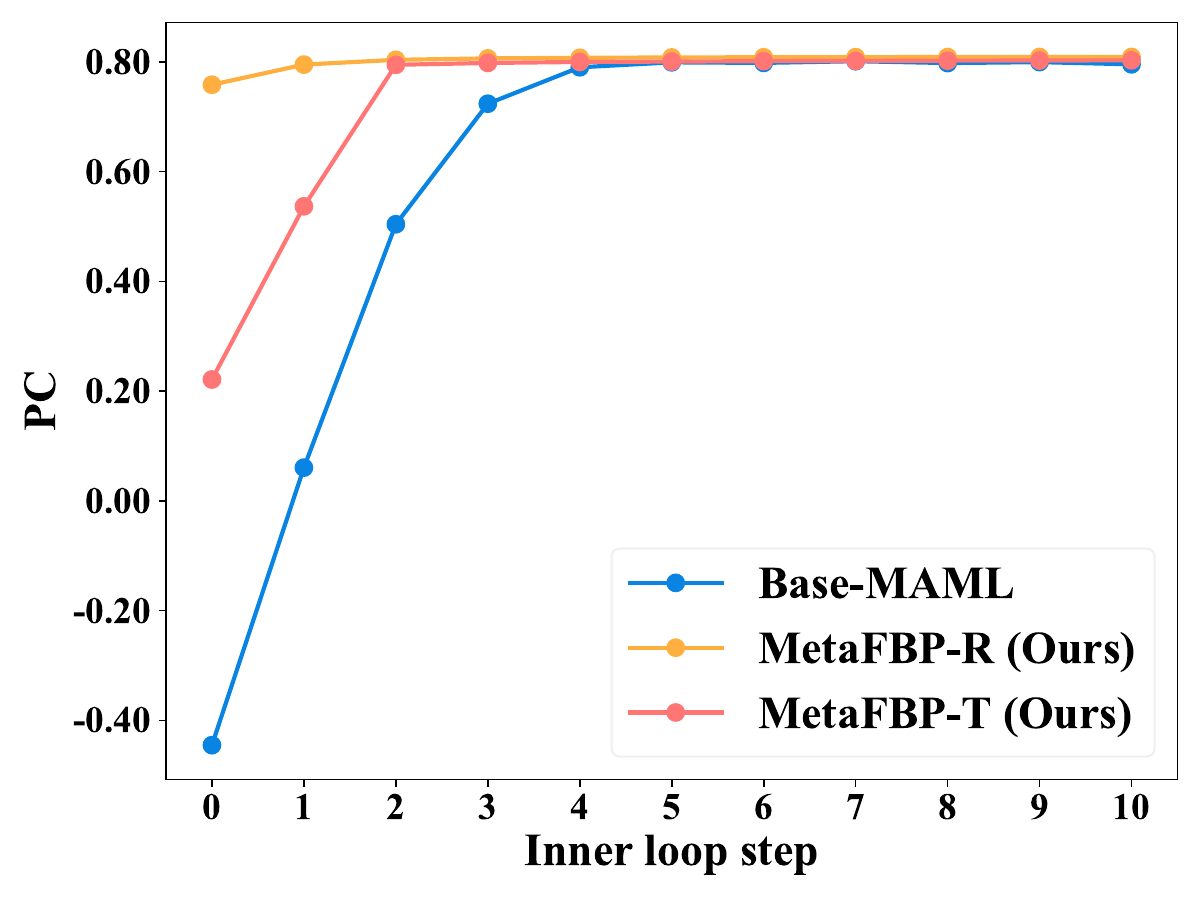}
         \vspace{-0.5cm}
         \caption{Train/Test with 15-shot.}
         \label{fig:mete_update_15shot}
    \end{subfigure}
    \vspace{-0.4cm}
        \caption{Pearson Correlation (PC) with respect to $k$-step in the inner loop of different models under different K-shot settings on PFBP-SCUT5500 benchmark.}
        \label{fig:mete_update}
        \vspace{-0.4cm}
\end{figure}
\begin{figure}[tt]
\centering
\includegraphics[width=0.82\columnwidth]{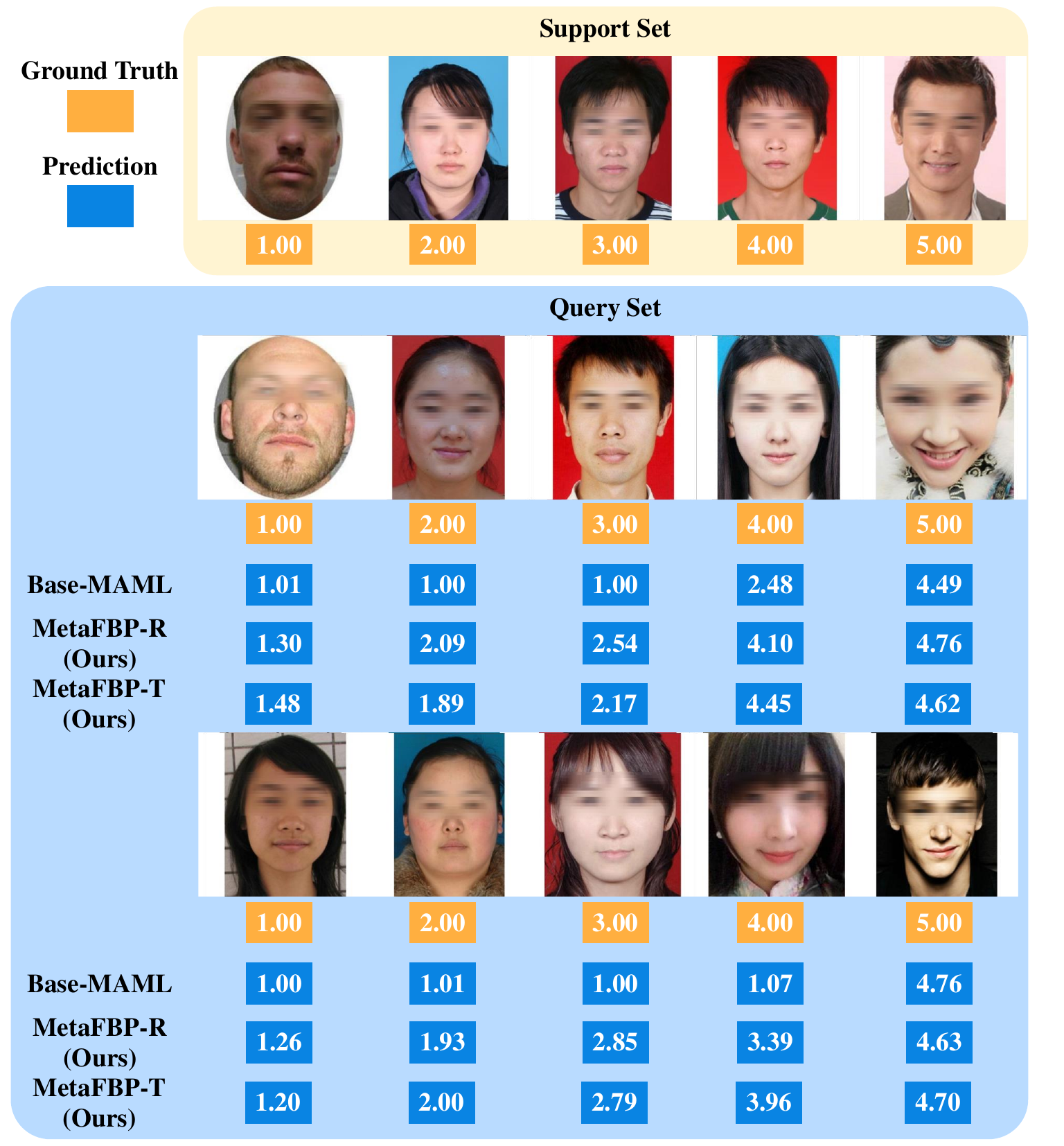}
 \vspace{-0.4cm}
\caption{The prediction results of a specific user provided by different models trained with PFBP-SCUT500 dataset .}
\label{fig:pred_vis}
 \vspace{-0.5cm}
\end{figure}

\subsection{Ablation Study}
\paragraph{\textbf{Different K-shot settings during training and testing phases.}} To further investigate the effectiveness of our method, we conduct extensive experiments on PFBP-SCUT5500, which train a model with specific shot number and test the model with different shot settings. We only report the results of PC, which are listed in Table \ref{tab:fbp5500_diffkshot}. We can find that for a model trained with specific shots, it can be improved with the increasing K-shot of support set. Our method (trained with 1-shot and tested with 1-shot) still outperforms the Base-MAML (trained with 1-shot and tested with 10-shot). It again demonstrates that our method has faster adaptation ability than MAML, even using less labeled data during fine-tuning.

\paragraph{\textbf{Exploring Adaptation Strength $\lambda$.}}
It shows in Equation \ref{eq:mix_parameres} that the adaptation strength $\lambda$ controls the adaptation magnitude. 
We investigate the effectiveness of different $\lambda$ on PFBP-SCUT5500 in Table \ref{tab:ablation_alpha}, from which we can observe that neither larger or smaller $\lambda$ can improve performance. Too large $\lambda$ may destroy the weights of the predictor so that causes drastic performance degradation. The smaller $\lambda$ can reduce the risk of over-fitting. However, too small $\lambda$ will make the high-order predictor finally trash into a plain predictor. Therefore, we set $\lambda$ to a normal value of 0.01.
\paragraph{\textbf{Visualization.}} An intuitive way to visualize the fast adaption of our method is shown in Figure \ref{fig:mete_update}. 
It can be seen that our method keeps the best performance with less variation on 1-shot (Figure \ref{fig:mete_update_1shot}) and 5-shot (Figure \ref{fig:mete_update_5shot}). For the 10-shot (Figure \ref{fig:mete_update_10shot}) and 15-shot (Figure \ref{fig:mete_update_15shot}) settings, our method earlier reaches the top max PC compared with MAML, which shows the proposed method can solve the slow adaptation and overfitting problems in conventional meta-learning methods. We also plot the prediction result for the most challenging 1-shot task on PFBP-SCUT500 benchmark. Figure \ref{fig:pred_vis} reveals that the Base-MAML model lacks ability to capture individual aesthetic preferences because it frequently assigns low scores to facial images, regardless of their actual differences. Conversely, our method can produce varying scores for different images, resulting in a higher correlation with the true labels. 

\section{Conclusion}
In this paper, we delve into Personalized Facial Beauty Prediction (PFBP). We model PFBP into a Few-Shot Learning (FSL) task and discuss its different challenge from conventional FSL task. We claim that PFBP requires a faster adaptation ability considering its user-adaptive characteristic, while the conventional meta-learning methods to solve FSL are usually trapped into slow adaptation or overfitting the tiny support set. To solve this problem, we develop a learning-to-learn mechanism into a high-order predictor for fast adaptation. Extensive quantitative and qualitative experiments demonstrate the effectiveness of the proposed method. 

\footnotesize{
\begin{acks}
This work was supported by the Fujian Provincial Natural Science Foundation (No. 2022J05135), the University-Industry Cooperation Project of Fujian Provincial Department of Science and Technology (No. 2020H6005), and the National Natural Science Foundation of China (No. U21A20471).
\end{acks}
}

\clearpage
\bibliographystyle{ACM-Reference-Format}
\balance
\bibliography{paper}

\end{document}